\documentclass[sigconf]{acmart}
\AtBeginDocument{%
  }

\usepackage{microtype}
\usepackage{graphicx}
\usepackage{subcaption}
\usepackage{multirow}
\usepackage{siunitx}
\usepackage{pifont}
\newcommand{\cmark}{\ding{51}}  
\newcommand{\xmark}{\ding{55}}  
\usepackage{booktabs}
\usepackage[most]{tcolorbox}
\usepackage[table]{xcolor}
\definecolor{other flat1}{RGB}{175,0,75}
\newcommand{\hi}[1]{\textbf{\textcolor{other flat1}{#1}}}
\usepackage{tikz}
\usepackage{balance}
\usetikzlibrary{shadows}

\usepackage{amsmath,amssymb,mathtools,amsthm}

\definecolor{promptbar}{RGB}{90,120,160}
\definecolor{promptbg}{RGB}{248,249,251}

\newtcolorbox{promptblock}[1][]{
  enhanced,
  breakable,
  colback=promptbg,
  colframe=promptbar,
  boxrule=0.5pt,
  arc=3pt,
  left=10pt,right=10pt,top=8pt,bottom=8pt,
  borderline west={3pt}{0pt}{promptbar},
  fonttitle=\bfseries\small,
  title={Prompt Template},
  coltitle=black,
  attach title to upper={\quad},
  #1
}

\usepackage{hyperref}
\usepackage[capitalize,noabbrev]{cleveref}

\copyrightyear{2026}
\acmYear{2026}
\setcopyright{cc}
\setcctype{by}
\acmConference[MM '26]{Proceedings of the 34th ACM International Conference on Multimedia}{November 10--14, 2026}{Rio de Janeiro, Brazil}
\acmBooktitle{Proceedings of the 34th ACM International Conference on Multimedia (MM '26), November 10--14, 2026, Rio de Janeiro, Brazil}
\acmDOI{10.1145/3767308.3836227}
\acmISBN{979-8-4007-2213-4/2026/11}

\begin{document}

\title{Beyond Visual Similarity: Entity-Aligned Retrieval for Knowledge-Based Visual Question Answering}

\author{Hangrui Xu}\authornotemark[1]
\affiliation{%
  \institution{Shenzhen International Graduate School, Tsinghua University}
  \city{Shenzhen}
  \country{China}}
\email{hangruixu666@gmail.com}

\author{Zhengxian Wu}\authornotemark[1]
\affiliation{%
  \institution{Shenzhen International Graduate School, Tsinghua University}
  \city{Shenzhen}
  \country{China}}
\email{zx-wu24@mails.tsinghua.edu.cn}

\author{Yunyao Yu}
\authornote{Equal contribution.}
\affiliation{%
  \institution{Shenzhen International Graduate School, Tsinghua University}
  \city{Shenzhen}
  \country{China}}
\email{yuyy25@mails.tsinghua.edu.cn}

\author{Zhuohong Chen}
\affiliation{%
  \institution{Shenzhen International Graduate School, Tsinghua University}
  \city{Shenzhen}
  \country{China}}
\email{zhuohong24@mails.tsinghua.edu.cn}

\author{Rui Cong}
\affiliation{%
  \institution{Shenzhen International Graduate School, Tsinghua University}
  \city{Shenzhen}
  \country{China}}
\email{congr24@mails.tsinghua.edu.cn}

\author{Xiangwen Deng}
\affiliation{%
  \institution{University of Arizona}
  \city{Tucson}
  \country{USA}}
\email{xiangwendeng@arizona.edu}

\author{Zhifang Liu}
\affiliation{%
  \institution{Shenzhen International Graduate School, Tsinghua University}
  \city{Shenzhen}
  \country{China}}
\email{liuzhifang@sz.tsinghua.edu.cn}

\author{Peng Jiao}
\affiliation{%
  \institution{Shenzhen International Graduate School, Tsinghua University}
  \city{Shenzhen}
  \country{China}}
\email{jiaop21@mails.tsinghua.edu.cn}

\author{Haoqian Wang}
\authornote{Corresponding author.}
\affiliation{%
  \institution{Shenzhen International Graduate School, Tsinghua University}
  \city{Shenzhen}
  \country{China}}
\email{wanghaoqian@tsinghua.edu.cn}

\renewcommand{\shortauthors}{Hangrui Xu et al.}

\begin{abstract}
Knowledge-Based Visual Question Answering (KB-VQA) relies on retrieving external information to answer queries involving long-tail entities. 
However, existing retrieval pipelines predominantly employ CLIP-style dual encoders, which prioritize surface-level visual similarity over entity-level semantic alignment. 
This paradigm often fails when semantically identical concepts exhibit large visual variations or when distinct entities appear visually similar. 
To address this, we propose KBMR, the first MLLM-based embedding retriever tailored for KB-VQA. 
Leveraging the robust autoregressive capabilities of MLLMs, KBMR maps images into a semantic space that better preserves concept identity. 
To tackle the challenge of noisy supervision in Wikipedia-scale retrieval, we introduce an MLLM-based semantic discriminator that generates continuous entity-consistency weights. 
These weights guide a novel continuous semantic distillation objective, enabling effective hard negative sampling and soft supervision beyond rigid binary labels. 
Extensive experiments demonstrate that KBMR significantly outperforms CLIP baselines, yielding up to a 14.7\% improvement in retrieval Recall@1 and a 9.4\% gain in end-to-end VQA accuracy. Code is available at \url{https://github.com/realHarryX/KBMR}.
\end{abstract}



\maketitle

\begin{figure}[!h]
\centering
\includegraphics[width=\linewidth]{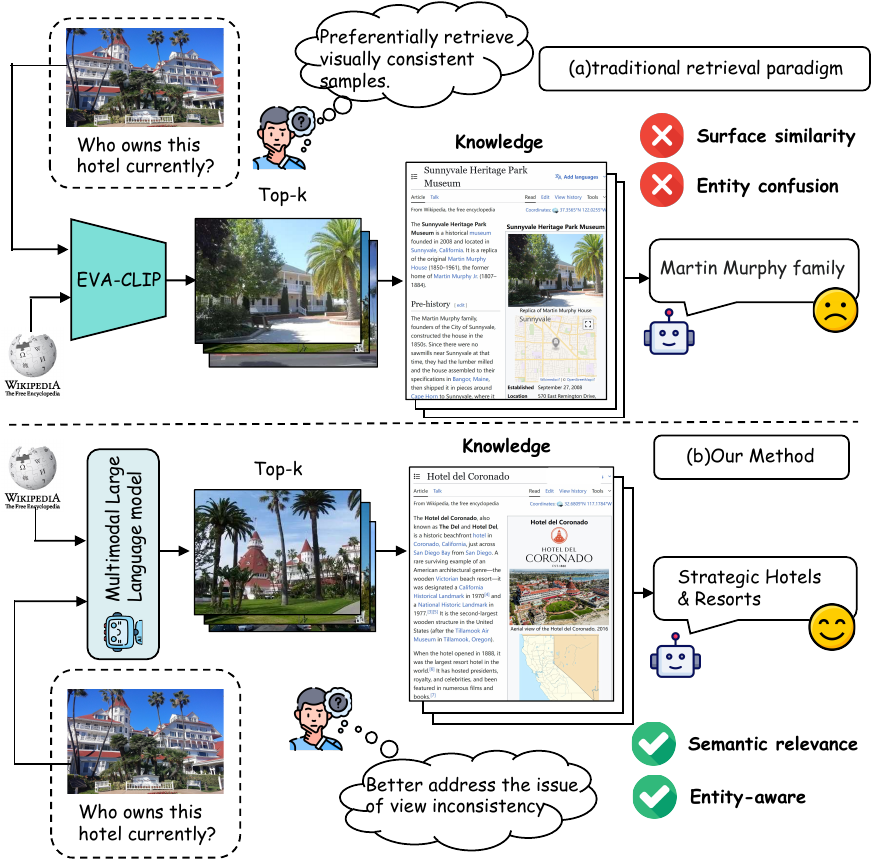}
\caption{\textbf{Motivation and overview of KBMR for KB-VQA retrieval.} Traditional CLIP retrieval prioritizes surface-level visual similarity, which is fragile to view inconsistency and often leads to entity confusion, recalling semantically mismatched evidence and producing incorrect answers. Our method leverages MLLM to retrieve entity-aware semantic knowledge entries, providing a higher-fidelity candidate pool and improving downstream answer grounding.}
\vspace{-4mm}
\label{fig:test1}
\end{figure}

\section{Introduction}

Multimodal large language models (MLLMs) have recently made remarkable progress in vision–language understanding and generation \cite{yu2026stabilizing}, enabling instruction following, multi-step reasoning \cite{li2026aim}, and open-ended responses over images.
However, many practical visual questions depend on external knowledge that is long-tailed and constantly evolving, where an MLLM’s parametric knowledge alone is unreliable and frequently leads to factual errors and hallucinated content \cite{lyu2026hallu_sae, lyu2025hallu_vdc, lyu2026hallu_pade, xiao2026staying}.
Knowledge-based visual question answering (KB-VQA) \cite{yan2024echosight, wu2026promsa, chen2026learning} addresses this limitation by augmenting MLLMs with Wikipedia-scale knowledge sources, retrieving relevant evidence and generating answers conditioned on it.

Most KB-VQA systems \cite{cocchi2025augmenting} follow a retrieval-augmented generation (RAG) pipeline that first retrieves candidate knowledge entries, then reranks or filters them, and finally generates an answer grounded in the most relevant evidence \cite{liu2026opera, liu2026faithfulness}.
This design makes first-stage retrieval a critical bottleneck.
In practice, existing KB-VQA systems \cite{yan2024echosight} almost universally adopt CLIP-style dual encoders as retrievers, mainly for their robust fine-grained visual representations and their vision–language alignment in a shared embedding space.
However, KB-VQA places greater emphasis on aligning entities under large appearance variations.
In Wikipedia corpora, the same concept can appear with substantial visual differences across viewpoints, time periods, and styles, while different entities can look highly similar (see Fig. \ref{fig:test1}).
Robustly matching the same subject across such viewpoint and appearance changes has long been a challenge for visual matching and recognition \cite{11595022, 10720843, 11112633, 11068977, fang2025cogstereoneuralstereomatching, xu2026psgait, wu2025dagait}.
Retrieval in KB-VQA is therefore inherently semantics-driven, and surface-level visual similarity does not necessarily align with actual relevance.
Prior work \cite{yang2025omgm} has explored external strategies such as multi-stage retrieval and query processing, but the candidate pool is still produced by a CLIP-style retriever, leaving the overall pipeline constrained by the CLIP paradigm.

In contrast, semantic autoregressive encoding in MLLMs \cite{jiang2025vlm2vectrainingvisionlanguagemodels} provides a more expressive representation for modeling fine-grained relations at the entity and concept levels.
Recent studies\cite{zhang2024gme, zhou2025megapairs, ConeSep, ENCODER, COMBINER} have begun to use MLLM embeddings as retrieval vectors for general cross-modal retrieval, targeting broad semantic relevance or cross-modal matching \cite{yang2026stable}.
Since their objectives align paired data across modalities, such as images and texts, the encoder mainly learns high-level semantics shared across modalities, such as object categories or scene concepts.
In KB-VQA, however, the goal is not to bridge modality gaps or return roughly relevant content, but to identify the knowledge evidence that precisely corresponds to the queried entity.
This requires the MLLM encoder to represent global semantics while also preserving fine-grained discriminative cues that separate visually similar entities.

To address these challenges, we propose \textbf{KBMR}, an MLLM-based retriever tailored for KB-VQA and, to the best of our knowledge, the \textbf{first} work to employ an MLLM as the retriever for this task.
KBMR trains an MLLM retriever whose image similarity is consistent with entity-level semantic relevance in knowledge retrieval.
Concretely, at the representation level, we prompt the MLLM to encode images into a shared semantic space and take the final-token hidden state as the retrieval embedding.
At the training sample level, we introduce an MLLM-based semantic discriminator that judges whether each query–candidate pair refers to the target entity, producing continuous entity-consistency weights.
These weights guide hard negative sampling to select high-quality and diverse negatives, and further serve as soft labels that relax the strict one-to-one mapping assumption.
Finally, we propose a continuous semantic distillation training strategy that aligns the retriever’s similarity distribution with the discriminator-induced weight distribution, encouraging finer-grained, entity-centric discrimination under long-tail knowledge and strong visual ambiguity.
Experiments on multiple benchmarks show that KBMR consistently improves retrieval recall over CLIP-family retrievers with a maximum gain of +14.7\% R@1, achieving state-of-the-art retrieval performance in KB-VQA.
The contributions of this work can be summarized as:
\begin{itemize}
    \item We propose \textbf{KBMR}, the first MLLM-based retriever for KB-VQA, which aligns retrieval similarity with entity-level semantic relevance under large appearance variations.
    \item We introduce an \textbf{MLLM-based semantic discriminator} to produce continuous \textbf{entity consistency weights}, and leverage them to enable robust hard-negative mining and soft supervision.
    \item We develop a \textbf{continuous semantic distillation} objective that aligns the retriever similarity distribution with weight distribution, encouraging entity-centric discrimination in highly confusing neighborhoods.
    \item Extensive experiments on multiple benchmarks show that KBMR achieves the best retrieval performance and delivers the strongest end-to-end KB-VQA accuracy.
\end{itemize}

\begin{figure*}[t]
\centering
\includegraphics[width=\linewidth]{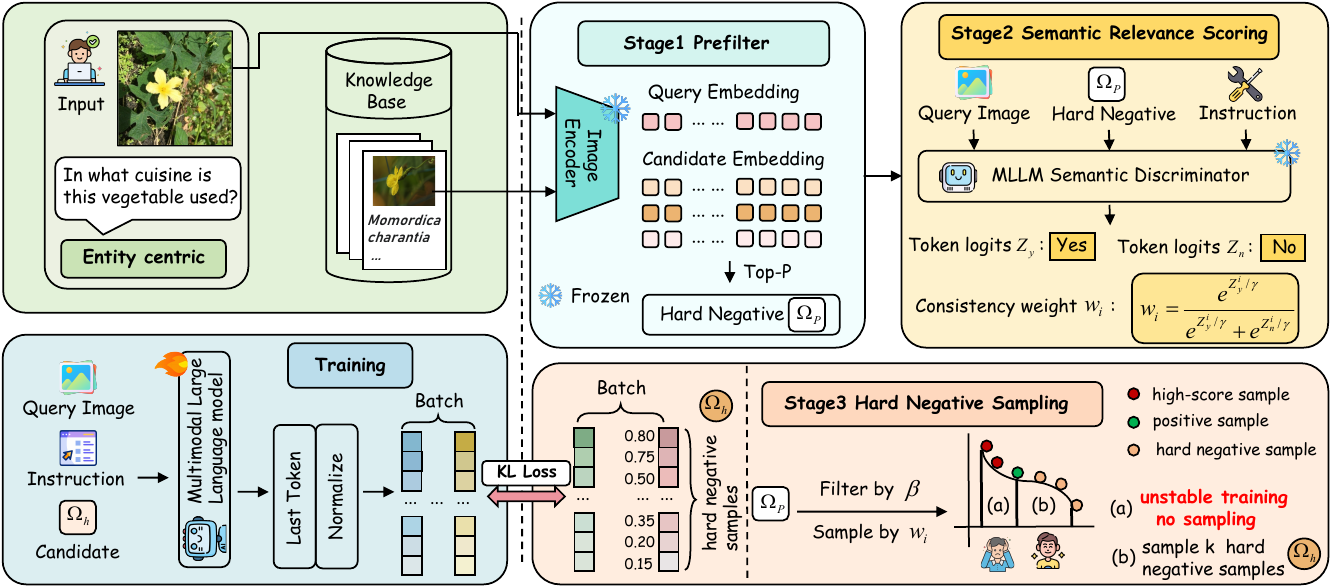}
\caption{\textbf{KBMR with hard negative sampling and continuous semantic distillation.}
Given an entity-centric KB-VQA query, Stage-1 uses a frozen encoder to retrieve a candidate pool. Stage-2 applies an MLLM-based Semantic Discriminator to score each query--candidate pair and produce entity consistency weights. Stage-3 samples hard negatives via weights to form training data. We train the retriever via continuous semantic distillation by aligning its similarity distribution with the discriminator-induced soft prior using a KL-based loss. \textbf{Right:} training data construction. \textbf{Left:} continuous semantic distillation training.}
\vspace{-2mm}
\label{fig:method}
\end{figure*}

\section{Related Work}
\subsection{Multimodal Large Language Models}
In recent years, multimodal large language models (MLLMs) have made substantial progress in both model architecture and capability \cite{bai2025qwen25vltechnicalreport, wu2026models, 11463453, gu2025unimev2mllmasajudgeuniversalmultimodal, wu2026language}.
Most existing approaches couple visual encoders with large language models through cross-modal alignment modules \cite{xiao2026promptbased}, thereby preserving strong language generation ability while significantly improving the understanding of complex visual content.
For instance, Qwen2.5-VL \cite{bai2025qwen25vltechnicalreport} strengthens high-resolution visual perception and multimodal reasoning, showing robust performance in challenging scenarios such as document understanding, chart analysis, and long-video modeling.
Benefiting from such perception and reasoning, MLLMs have been widely adapted to downstream tasks, ranging from fine-grained perception and quality assessment \cite{zhong2025adaptive, zhong2024causal, zhong2025semi} to higher-level semantic understanding such as sarcasm interpretation \cite{wang2025can}, emotion understanding \cite{fang2026emoagentr1multimodalemotionunderstanding}, consistency-based forgery reasoning \cite{wang2026lapforensics}, and human motion–language grounding \cite{lyu2025towards}.
Architecturally, recent research has moved from loosely connected multimodal components toward deeper cross-modal fusion with joint pretraining \cite{xiao2026layer, xiao2025visual}.
On the one hand, native multimodal Transformers let different modalities interact directly within a shared representation space, echoing the benefit of structured and fine-grained representation learning observed in dense visual understanding tasks \cite{11208933}.
On the other hand, multimodal pretraining now goes beyond image–text alignment to include cross-modal generation \cite{yuan2025dissecting, meng2026physical}, temporal understanding, and multimodal instruction following.
In parallel, multimodal generative models have advanced in controllable image and video synthesis and editing \cite{zhong2025ctd, ma2026group, wang2026liveedit}, with growing attention to physical plausibility \cite{meng2025phymagic, meng2025grounding} and inference efficiency \cite{zheng2025compute, zhao2026resilphase, zhao2026seeingendstepzero}.

\subsection{Knowledge-based Visual Question Answering}
KB-VQA focuses on questions that require external knowledge to answer \cite{hong2025knowledge, chen2026learning}, where the needed information is often long-tailed and fine-grained and cannot be inferred from the image alone.
To bridge this gap, recent work \cite{yan2024echosight, wu2026promsa} commonly formulates KB-VQA as a multimodal retrieval-augmented generation (RAG) problem: the system first retrieves relevant evidence from large-scale knowledge sources, such as Wikipedia articles and associated images, and then conditions a vision–language model on the retrieved content to better handle knowledge-sparse and long-tail queries \cite{li2026ragtrack}.

Existing KB-VQA systems follow the CLIP paradigm in first-stage retrieval \cite{ling2025mmkb, yang2025omgm, zhang2024mr}, encoding the query image and each candidate knowledge-item image separately and constructing a candidate set by vector similarity \cite{yang2026stable}.
Building upon this paradigm, prior work has mainly improved retrieval quality through external enhancements rather than fundamentally revisiting the retriever itself. One line of research learns stronger multimodal representations \cite{li2026cadtrack} via improved image--text alignment or better joint embedding spaces \cite{ConeSep, ENCODER, COMBINER}, to retrieve more relevant entities and evidence.
Another uses post-retrieval re-ranking, filtering, or query refinement \cite{cocchi2025augmenting, caffagni2024wiki, hong2025knowledge, liu2026opera}.
For example, Wiki-LLaVA \cite{caffagni2024wiki} employs a hierarchical pipeline that first retrieves entities/documents then drills down to passages; ReflectiVA \cite{cocchi2025augmenting} uses self-reflective tokens for an MLLM to judge retrieval necessity and assess passage relevance for filtering; and Wiki-PRF \cite{hong2025knowledge} refines search via tool-driven query augmentation (e.g., flipping) and aggregation, with filtering to boost evidence quality.
Beyond retrieval itself, another line improves how retrieved evidence is exploited during answer generation, either by reasoning more faithfully over it \cite{liu2026faithfulness, li2026aim} or by suppressing hallucinated statements inconsistent with it \cite{lyu2026hallu_sae, lyu2025hallu_vdc}.

While these methods help, they still operate on candidate sets generated by a CLIP-style retriever.
Consequently, retrieval remains constrained by CLIP’s representation space and similarity metric \cite{xiao2026not}, which emphasize visual similarity and often miss the fine-grained, entity-centric semantic relevance required for KB-VQA, especially under long-tail settings \cite{li2025chi, li2026cpgrec+, li2024category}.
In contrast, rather than stacking refinement modules on top of the CLIP paradigm, we \textbf{break this limitation} at the retriever level by employing an MLLM as the retriever, leveraging autoregressive semantic representations to model concept-level alignment for KB-VQA.

\section{Methodology}

\subsection{Overview}
Knowledge-Based Visual Question Answering (KB-VQA) requires models to answer questions about input images, often involving long-tail entities, by leveraging an external knowledge base.
The standard pipeline leverages the input image to query a massive corpus for images of the same entity, thereby accessing its associated knowledge.
However, this process is fundamentally limited by the gap between superficial visual similarity and genuine semantic relevance.
We propose \textbf{KBMR}, an MLLM-based embedding retriever (\S\ref{sec:mllm_retriever}) that changes the embedding space from pixel-based matching to an entity-aligned semantic space (see Fig. \ref{fig:method}).
At its core is a \textbf{Semantic Discriminator (SD)} (\S\ref{sec:semantic_discriminator}) that scores each query--candidate pair and yields \textbf{Entity Consistency Weights}, which in turn guide hard negative sampling (\S\ref{sec:hard_negative_mining}) to pick high-quality and diverse hard negatives for training.
KBMR is then trained with \textbf{Continuous Semantic Distillation} (\S\ref{sec:continuous_distillation}), which pulls the retriever's similarity distribution toward the SD-based semantic prior by minimizing the discrepancy between the two distributions.
At inference, the trained retriever performs fast similarity search in this refined space to recall high-quality candidates.
Since we rely on distribution alignment instead of hard binary labels, KBMR builds an entity-level structure in the latent space and improves retrieval precision under long-tail knowledge and strong visual ambiguity.

\subsection{MLLM-based Embedding Retriever}\label{sec:mllm_retriever}
Unlike the dual-tower structure of CLIP, MLLM incorporates three components: a vision tower, a projection layer, and an LLM backbone.
While preserving fine-grained visual details, MLLMs unify image content with language instructions into an autoregressive semantic space, producing context-aware, concept-level representations.
Inspired by the field of embedding~ \cite{jiang2024e5}, we employ the prompt: ``\texttt{\textless Image\textgreater\ Summary above image in one word:\textbackslash n}'' to instruct the MLLM to summarize image into a shared semantic space, and take the last-token hidden state as the embedding, which yields image embeddings with a low modality gap and rich semantics.

\subsection{Semantic Discriminator as Soft Supervision}\label{sec:semantic_discriminator}
To begin with, we detail the construction of the training data for the embedding retriever.
Since KB-VQA retrieval targets entity-level relevance rather than surface visual similarity, we introduce an MLLM as \textbf{Semantic Discriminator (SD)} to produce an \textbf{entity consistency weight} that guides hard negatives sampling and yields reliable, diverse hard evidence.

Given the input query and the candidate sample set (defined in Section \ref{sec:hard_negative_mining}), the SD computes the entity consistency weight for each query-candidate pair under the following instruction: ``\texttt{You need to determine whether the given Candidate and the Query refer to the same entity. If they do, answer "Yes"; otherwise, answer "No". Query:$<Query>$, Candidates:$<Candidate>$.}''
We then compute the entity consistency weight $\mathrm{W}=\{w_1,w_2,\dots,w_m\}$ from the logits of the \texttt{Yes}~($z_y$) and \texttt{No}~($z_n$) tokens using calibrated logit scaling:
\begin{equation}
w_i=\frac{exp(z_y^i/\gamma)}{exp(z_y^i/\gamma) + exp(z_n^i/\gamma)} = \sigma\!\left(\frac{z_y^i-z_n^i}{\gamma}\right)
\end{equation}
where $\sigma(\cdot)$ denotes the sigmoid function and $\gamma>0$ is a semantic sharpness coefficient controlling weight smoothness (larger $\gamma$ produces softer weights). 
Here, $\mathrm{W}\in\mathbb{R}^{n_q\times 50}$ and $n_q$ denotes the number of queries. 
Leveraging MLLMs' advanced understanding capabilities, the entity consistency weight $\mathrm{W}$ effectively captures the degree of semantic alignment between queries and candidates.

\sisetup{
  detect-weight=true,
  detect-inline-weight=math
}

\subsection{Hard Negative Sampling}\label{sec:hard_negative_mining}
This section describes how we mine hard negative samples suitable for training from the large-scale retrieval corpus.
\noindent{\textbf{Potential Hard Negative Set.}} 
We first use EVA‑CLIP to generate embeddings for the query image and all database images.
After excluding all positive samples, we retrieve the top‑50 most similar candidates for each query to form a potential hard negative set $\Omega_p$:
\begin{equation}
\begin{aligned}
    \Omega_{p} = \text{Rank}_{50}(cos(e_{q}, e_{c})), \text{where}\ e_{c} \neq e^{+}_{c},
\end{aligned}
\end{equation}
where $e^{+}_{c}$ denote positive candidate, $e_{q}$ is the query embedding, and $e_{c}$ represents all candidate embeddings. The function $cos(\cdot)$ computes pairwise similarity scores, and $\text{Rank}_{k}$ selects the top-$k$ highest-scoring candidates as potential hard negatives.

To improve the quality of hard negatives, we use the Semantic Discriminator described in \ref{sec:semantic_discriminator} to compute the entity consistency weight $\mathrm{W}$ between the query and each candidate, as well as between the query and the positive sample.
Based on the query--positive weight $w_{q,c_t}$, we define a threshold $\alpha = w_{q,c_t} - \beta$, where $\beta$ is a hyper‑parameter controlling the margin.
We then remove from $\Omega_p$ all candidates whose entity consistency weight exceeds $\alpha$, filtering out samples that remain semantically highly relevant to the query and could cause confusion.
It is worth emphasizing that candidates with high embedding similarity but low entity consistency are particularly informative: they lie close to the query in the current representation space but are semantically mismatched.

To ensure diversity in difficulty, we perform stratified sampling according to the entity consistency weights: the remaining candidates are divided into four difficulty strata based on their weights, and two samples are randomly drawn from each stratum to form a fixed‑size hard negative set.
If the resulting set contains fewer than eight samples, we duplicate selections to guarantee a minimum of eight; in the rare case where no candidate meets the criteria, the training sample is discarded.
Finally, for each query $q$, we obtain a hard negative set $\Omega_h = {c_1, \dots, c_k}$ together with the corresponding entity consistency weights $\mathrm{W_h}=\{w_{q,c_1},...,w_{q,c_k}\}$.

\begin{table*}[t]
  \centering
  \setlength{\tabcolsep}{.4em}
  \caption{\textbf{VQA accuracy on E-VQA and InfoSeek.} $\dagger$ indicates the EchoSight model without reranking, and $\ddagger$ indicates the EchoSight model with reranking. $\diamond$ indicates that MMKB-RAG is not open source, so we obtained the results by replicating the paper. Rows in blue correspond to replacing the retriever with KBMR. Red values denote absolute performance improvements.}
  \vspace{-2mm}
  \begin{tabular}{c|c|S[table-format=2.1]S[table-format=2.1]S[table-format=2.1]S[table-format=2.1]S[table-format=2.1]}
   \toprule
    \multirow{2.5}{*}{\textbf{Method}} & \multirow{2.5}{*}{\textbf{w/ KBMR}} & \multicolumn{2}{c}{\textbf{E-VQA}} & \multicolumn{3}{c}{\textbf{InfoSeek}} \\
    \cmidrule(lr){3-4} \cmidrule(lr){5-7}
    & & \multicolumn{1}{c}{Single-Hop} & \multicolumn{1}{c}{All} & \multicolumn{1}{c}{Unseen-Q} & \multicolumn{1}{c}{Unseen-E} & \multicolumn{1}{c}{All} \\
    \midrule

    \multirow{3}{*}{EchoSight\textsuperscript{$\dagger$} \cite{yan2024echosight}}
    & \xmark & 19.1 & 19.4 & 27.1 & 27.9 & 27.7 \\
    & \cellcolor{cyan!15}\cmark
    & \cellcolor{cyan!15}26.7
    & \cellcolor{cyan!15}27.6 
    & \cellcolor{cyan!15}35.1
    & \cellcolor{cyan!15}36.3
    & \cellcolor{cyan!15}36.0 \\
    & $\Delta$
    & \multicolumn{1}{c}{\hi{+7.6}}
    & \multicolumn{1}{c}{\hi{+8.2}}
    & \multicolumn{1}{c}{\hi{+8.0}}
    & \multicolumn{1}{c}{\hi{+8.4}}
    & \multicolumn{1}{c}{\hi{+8.3}} \\
    \midrule

    \multirow{3}{*}{EchoSight\textsuperscript{$\ddagger$} \cite{yan2024echosight}}
    & \xmark & 42.0 & 41.8 & 30.7 & 31.5 & 31.3 \\
    & \cellcolor{cyan!15}\cmark
    & \cellcolor{cyan!15}51.3
    & \cellcolor{cyan!15}51.0 
    & \cellcolor{cyan!15}38.7
    & \cellcolor{cyan!15}39.6
    & \cellcolor{cyan!15}39.2 \\
    & $\Delta$
    & \multicolumn{1}{c}{\hi{+9.3}}
    & \multicolumn{1}{c}{\hi{+9.2}}
    & \multicolumn{1}{c}{\hi{+8.0}}
    & \multicolumn{1}{c}{\hi{+8.1}}
    & \multicolumn{1}{c}{\hi{+7.9}} \\
    \midrule

    \multirow{3}{*}{ReflectiVA \cite{cocchi2025augmenting}}
    & \xmark & 28.0 & 29.2 & 40.4 & 39.8 & 40.1 \\
    & \cellcolor{cyan!15}\cmark
    & \cellcolor{cyan!15}34.7
    & \cellcolor{cyan!15}35.2 
    & \cellcolor{cyan!15}48.5
    & \cellcolor{cyan!15}48.9
    & \cellcolor{cyan!15}48.6 \\
    & $\Delta$
    & \multicolumn{1}{c}{\hi{+6.7}}
    & \multicolumn{1}{c}{\hi{+6.0}}
    & \multicolumn{1}{c}{\hi{+8.5}}
    & \multicolumn{1}{c}{\hi{+8.1}}
    & \multicolumn{1}{c}{\hi{+8.5}} \\
    \midrule

    \multirow{3}{*}{MMKB-RAG\textsuperscript{$\diamond$} \cite{ling2025mmkb}}
    & \xmark & 39.7 & 35.9 & 36.4 & 36.3 & 36.4 \\
    & \cellcolor{cyan!15}\cmark
    & \cellcolor{cyan!15}47.3
    & \cellcolor{cyan!15}43.1 
    & \cellcolor{cyan!15}45.6
    & \cellcolor{cyan!15}45.7
    & \cellcolor{cyan!15}45.7 \\
    & $\Delta$
    & \multicolumn{1}{c}{\hi{+7.6}}
    & \multicolumn{1}{c}{\hi{+7.2}}
    & \multicolumn{1}{c}{\hi{+9.2}}
    & \multicolumn{1}{c}{\hi{+9.4}}
    & \multicolumn{1}{c}{\hi{+9.3}} \\
    \midrule

    \multirow{3}{*}{OMGM \cite{yang2025omgm}}
    & \xmark & 49.7 & 50.2 & 43.5 & 43.5 & 43.5 \\
    & \cellcolor{cyan!15}\cmark
    & \cellcolor{cyan!15}{\textbf{55.0}}
    & \cellcolor{cyan!15}{\textbf{54.7}} 
    & \cellcolor{cyan!15}{\textbf{50.9}}
    & \cellcolor{cyan!15}{\textbf{50.8}}
    & \cellcolor{cyan!15}{\textbf{50.8}} \\
    & $\Delta$
    & \multicolumn{1}{c}{\hi{+5.3}}
    & \multicolumn{1}{c}{\hi{+4.5}}
    & \multicolumn{1}{c}{\hi{+7.4}}
    & \multicolumn{1}{c}{\hi{+7.3}}
    & \multicolumn{1}{c}{\hi{+7.3}} \\
    
  \bottomrule
  \end{tabular}
\label{tab:main2}
\vspace{-2mm}
\end{table*}

\subsection{Continuous Semantic Distillation Training}\label{sec:continuous_distillation}

Standard contrastive learning enforces a rigid one-to-one matching objective, treating the target candidate as the only positive and all remaining candidates uniformly as negatives. Although effective for generic visual-text alignment, this formulation is suboptimal for KB-VQA retrieval, where hard negatives are not equally irrelevant: some candidates are visually similar yet semantically mismatched, while others still share partial semantic relevance with the query entity. One-hot supervision therefore cannot adequately reflect the fine-grained semantic structure within a highly confusable candidate neighborhood.

To address this issue, we cast training as a \emph{semantic distribution distillation} problem. Instead of separating positives from negatives using discrete labels, we encourage the retriever to match a soft entity-aware target distribution induced by the SD, so that it not only favors the target candidate but also respects the semantic relationships among candidates in the neighborhood. This provides richer supervision than standard contrastive learning and is beneficial for entity-centric retrieval under strong visual ambiguity.

Given a query $q$ and its candidate set $\Omega_c=\{c_t,c_1,\dots,c_k\}$,
where $c_t$ denotes the target candidate and $\{c_1,\dots,c_k\}$ are the hard negatives, we feed the query and all candidates into the trainable MLLM retriever and extract the last-token hidden states as embeddings. 
Let $e_q$ denote the query embedding, and let $E_c=\{e_c^{+},e_{c_1}^{-},\dots,e_{c_k}^{-}\}$ denote the candidate embeddings, where $e_c^{+}$ corresponds to the target candidate and $e_{c_i}^{-}$ corresponds to the $i$-th hard negative.

We first define a query-conditioned retriever posterior over the candidate set by normalizing the retrieval similarities with a temperature-scaled softmax:
\begin{equation}
\mathbf{p}_q(j)=
\frac{\exp\!\left(\mathrm{sim}(e_q,e_j)/\tau\right)}
{\sum\limits_{e \in E_c}\exp\!\left(\mathrm{sim}(e_q,e)/\tau\right)},
\quad e_j \in E_c,
\end{equation}
where $\mathrm{sim}(\cdot,\cdot)$ denotes cosine similarity and $\tau$ is the retrieval temperature. Intuitively, $\mathbf{p}_q(j)$ measures how much probability mass the current retriever assigns to the $j$-th candidate under the query-conditioned retrieval space, and collectively $\mathbf{p}_q=\{\mathbf{p}_q(j)\}_{j\in\{t,1,\dots,k\}}$ reflects the retriever's current belief about candidate relevance.

Meanwhile, the semantic discriminator provides entity consistency weights $W_c=\{w_{q,c_t},w_{q,c_1},\dots,w_{q,c_k}\}$,
where each weight indicates the degree of semantic alignment between the query and a candidate at the entity level. Rather than treating these weights as independent scalar scores, we further transform them into a normalized soft semantic prior over the same candidate set:
\begin{equation}
\mathbf{s}_q(j)=
\frac{\exp\!\left(w_{q,c_j}/\tau\right)}
{\sum\limits_{c \in \Omega_c}\exp\!\left(w_{q,c}/\tau\right)},
\quad c_j \in \Omega_c,
\end{equation}
where $\tau$ controls the sharpness of the semantic prior. Compared with a one-hot target, $\mathbf{s}_q$ preserves the relative semantic structure among candidates: those with higher semantic consistency receive larger prior mass, while semantically mismatched ones are suppressed. $\mathbf{s}_q$ can thus be interpreted as a soft teacher distribution that provides richer supervision for retrieval learning.

Based on the two distributions defined on the same candidate set, we train the retriever by minimizing the discrepancy between the retriever posterior $\mathbf{p}_q$ and the semantic prior $\mathbf{s}_q$ via symmetric KL divergence:
\begin{equation}
\mathcal{L}_{\mathrm{CSD}}
=
\mathbb{E}_{q \sim \mathcal{D}}
\left[
\frac{1}{2}
\Big(
D_{\mathrm{KL}}(\mathbf{p}_q \,\|\, \mathbf{s}_q)
+
D_{\mathrm{KL}}(\mathbf{s}_q \,\|\, \mathbf{p}_q)
\Big)
\right].
\end{equation}

This objective serves two purposes. The term $D_{\mathrm{KL}}(\mathbf{p}_q \,\|\, \mathbf{s}_q)$ encourages the retriever to move its probability mass toward semantically correct candidates, while the reverse term $D_{\mathrm{KL}}(\mathbf{s}_q \,\|\, \mathbf{p}_q)$ prevents the learned distribution from collapsing into an overly sharp or biased posterior and improves optimization stability. The retriever thus learns a candidate distribution that is both discriminative with respect to the target candidate and consistent with the semantic structure estimated by the discriminator.

In essence, our continuous semantic distillation replaces rigid discrete supervision with distribution-level semantic supervision over the entire hard-negative neighborhood. By matching the retriever posterior with the entity-aware semantic prior, the model learns a more structured and semantically calibrated embedding space, leading to stronger entity discrimination and more reliable retrieval for KB-VQA.

\begin{table*}[t]
  \centering
  \vspace{-1mm}
  \setlength{\tabcolsep}{.4em}
  \caption{\textbf{Retrieval performance on E-VQA and InfoSeek.}  \textsuperscript{$\blacktriangle$} denotes the universal retrieval methods.}
  \vspace{-2mm}
  \begin{tabular}{lc cccc c cccc}
   \toprule
    \multirow{2.5}{*}{\textbf{Retriever}} &  \multicolumn{4}{c}{\textbf{E-VQA}} & & \multicolumn{4}{c}{\textbf{InfoSeek}} \\
    \cmidrule{2-5} \cmidrule{7-10}
    & R@1 & R@5 & R@10 & R@20 &  & R@1 & R@5 & R@10 & R@20 \\
    \midrule
    \rowcolor{lightgray} 
    \multicolumn{10}{l}{\textit{CLIPs}} \\
    CLIP I-T \cite{clip}                        & 3.3  & 7.7  & 12.1 & 16.5 &  & 32.0 & 54.0 & 61.6 & 68.2 \\
    CLIP ViT-L/14 \cite{clip}                   & 9.9  & 22.0 & 27.8 & 31.7 &  & 22.5 & 40.4 & 42.7 & 44.1 \\
    EVA-CLIP-8B \cite{sun2024eva}               & 13.3 & 31.3 & 41.0 & 48.8 &  & 45.6 & 67.1 & 73.0 & 77.9 \\
    \midrule
    \rowcolor{lightgray} 
    \multicolumn{10}{l}{\textit{MLLM Retriever}} \\
    BLIP-2 \cite{li2023blip2}                   & 4.9  & 9.7 & 15.7 & 20.6 &  & 10.8 & 18.7 & 21.4 & 23.9 \\
    Qwen2-VL-7B \cite{wang2024qwen2vl}          & 6.1 & 14.7 & 18.5 & 21.9 &  & 17.5 & 31.4 & 35.1 & 37.6 \\
    LLaVA-OneVision \cite{li2024llava} & 7.4 & 15.1 & 23.9 & 29.5 &  & 21.8 & 42.6 & 44.7 & 47.2 \\
    GME\textsuperscript{$\blacktriangle$} \cite{zhang2024gme}
                                               & 10.7 & 24.1 & 33.4 & 44.6 &  & 38.7 & 59.9 & 63.7 & 67.8 \\
    MM-Ret\textsuperscript{$\blacktriangle$} \cite{zhou2025megapairs}            
                                               & 11.9 & 26.4 & 37.1 & 49.6 &  & 41.3 & 63.3 & 67.5 & 71.8 \\
    \midrule
    \rowcolor{lightgray} 
    \multicolumn{10}{l}{\textit{MLLMs with Continuous Semantic Distillation Training}} \\
    \rowcolor{cyan!15}
    \textbf{KBMR} (Qwen2-VL-7B) & 22.6 & 43.1 & 50.2 & 53.5 &  & 55.8 & 71.1 & 77.4 & 83.2 \\
    \rowcolor{cyan!15}
    \textbf{KBMR} (LLaVA-OV-7B) & \textbf{24.7} & \textbf{48.7} & \textbf{52.4} & \textbf{55.0} &  & \textbf{60.3} & \textbf{74.7} & \textbf{79.6} & \textbf{84.4} \\
  \bottomrule
  \end{tabular}
\label{tab:main1}
\vspace{-2mm}
\end{table*}

\section{Experiments}

\subsection{Experiment Setup}
\textbf{Datasets and Metrics.}
We evaluate on three KB-VQA benchmarks: Encyclopedic VQA (E-VQA) \cite{enc-vqa}, InfoSeek~\cite{infoseek} and Outside Knowledge VQA (OK-VQA) \cite{marino2019okvqa}.
E-VQA asks fine-grained questions about natural species and landmarks \cite{van2021ina_benchmarking, weyand2020google_landmark}; InfoSeek poses information-seeking questions over Wikipedia entities from OVEN \cite{hu2023open}, with its validation split disjoint from training in both entities and questions; OK-VQA requires knowledge beyond MS COCO images \cite{lin2014microsoft} across diverse domains.
Following previous setting, we report results on the E-VQA test set and on the entire InfoSeek validation split with a knowledge base of 100K Wikipedia entries.
For each benchmark we assess both retrieval and QA quality: retrieval is measured by recall, i.e., whether the correct article appears among the top-k retrieved results, while QA follows the original dataset protocol, using BEM score \cite{bulian2022tomayto} for E-VQA and VQA accuracy \cite{goyal2017making} for InfoSeek.

\textbf{Baselines.} 
We benchmark three retriever types: standard CLIPs visual encoder, MLLMs that directly output embeddings for similarity matching, and our trained MLLM retriever (KBMR). To verify end-to-end impact, we plug KBMR into a state-of-the-art KB-VQA pipeline by replacing only the retriever and keeping all other settings fixed, so improvements are solely due to better retrieval.

\textbf{Implementation Details.}
We first employ EVA-CLIP-8B to extract query and candidate embeddings, from which we construct a pool of potential hard negatives, and then utilize Qwen2.5-VL-7B to produce soft entity consistency weights for query–candidate pairs. We train KBMR using two MLLMs, Qwen2-VL~ \cite{wang2024qwen2vl} and LLaVA-OneVision~ \cite{li2024llava}, respectively.
To improve training efficiency and reduce GPU memory consumption, we adopt LoRA with rank 16 in conjunction with DeepSpeed ZeRO stage-2. All experiments are conducted on 8$\times$ NVIDIA A100 (80GB) GPUs. We resize all images to a resolution of 336$\times$336 and set the accumulated batch size to 1024. The learning rates are 1e-4 for Qwen2-VL and 2e-5 for LLaVA-OneVision. The model is trained for 5,000 steps on 600k samples drawn from the training sets of E‑VQA, InfoSeek, and OK‑VQA, with the hard negative count $k$ set to 8.

\subsection{Main Results}
\textbf{VQA Results.}
Tab.~\ref{tab:main2} reports end-to-end KB-VQA accuracy on E-VQA and InfoSeek, where the blue rows replace the original retriever with KBMR while keeping the pipeline unchanged (for OMGM, we replace its first-stage retrieval results with ours).
On E-VQA, KBMR brings gains of up to \textbf{+9.2}, which remain substantial even with a reranker (e.g., EchoSight‡), showing that upstream candidates complement later-stage refinement. On InfoSeek, KBMR improves performance by up to \textbf{+9.3}, demonstrating robustness to unseen questions/entities at Wikipedia scale. Our method establishes \textbf{new SOTA VQA accuracy of 54.7 on E-VQA and 50.8 on InfoSeek}.
These improvements are consistent across methods, indicating that KBMR benefits diverse downstream architectures by improving the retrieved candidate pool. Moreover, substantial gains for systems with reranking or stronger reasoning modules confirm that better first-stage retrieval complements rather than duplicates later-stage refinement. Overall, KBMR translates improved retrieval into reliable end-to-end VQA gains across diverse pipelines.

\textbf{Retrieval Results.}
Tab.~\ref{tab:main1} reports retrieval performance on E-VQA and InfoSeek. CLIP-style retrievers provide strong visual baselines, but are limited by the mismatch between visual similarity and entity-level semantic relevance. Although MLLMs are in principle better suited to entity-aware semantic matching, existing zero-shot MLLM embeddings often underperform strong CLIP retrievers, showing that general multimodal understanding does not directly yield an effective KB-VQA retrieval space. Similarly, state-of-the-art universal retrievers focus on broad multimodal alignment and may fail to capture the fine-grained entity-centric relations required by KB-VQA.
On E-VQA, KBMR (LLaVA-OV-7B) outperforms the best MLLM retriever baseline by \textbf{+12.8} points at R@1 and the strongest CLIP retriever (EVA-CLIP-8B) by \textbf{+11.4}. On InfoSeek, KBMR surpasses the strongest CLIP retriever by \textbf{+14.7} points at R@1 and the best MLLM retriever by \textbf{+19.0}. These consistent gains show that KBMR effectively converts MLLMs into high-recall KB-VQA retrievers by modeling entity-aware semantic relations, alleviating the candidate-set bottleneck and providing stronger evidence for downstream reasoning and answer grounding.

\begin{table}[ht]
\centering
\caption{Comparison of the retrieval and VQA results on OK-VQA. \textsuperscript{*}Results obtained by replacing OMGM's retriever.}
\vspace{-2mm}
\begin{tabular}{c | c c }
\toprule
Method & Pseudo Recall@5 & VQA score \\
\midrule
PreFLMR \cite{lin2024preflmr} & 70.9 & 61.9 \\
OMGM \cite{yang2025omgm} & 73.4  & 66.6  \\
Wiki-PRF-7B \cite{hong2025knowledge} & - & 77.8 \\
 \rowcolor{cyan!15}
KBMR(Qwen2-VL-7B) & 78.1  & 76.3\textsuperscript{*}  \\
 \rowcolor{cyan!15}
KBMR(LLaVA-OV-7B) & \textbf{80.4}  & \textbf{79.3\textsuperscript{*}}  \\
\bottomrule
\end{tabular}
\vspace{-2mm}
\label{tab:okvqa_eval}
\end{table}

\textbf{Results on more benchmarks.}
As shown in Tab.~\ref{tab:okvqa_eval}, we evaluate our model on the widely used OK-VQA benchmark. 
The two columns in the table report the retrieval accuracy and VQA accuracy of each method, respectively. 
The VQA accuracy of our method was tested by plugging our first-stage retrieval results into the OMGM pipeline while keeping other components unchanged. 
Compared to the original OMGM, our approach improves the answer accuracy by \textbf{+12.7} points. 
As shown in the table, KBMR significantly enhances retrieval performance and translates this gain into a substantial improvement in final answer accuracy, establishing a \textbf{new SOTA score of 79.3 on OK-VQA}.

\subsection{Ablation Studies}
\textbf{Are SD supervision signals truly effective for retrieval learning?}
To evaluate the effectiveness of SD supervision, we vary both the hard-negative construction strategy and training weights while fixing the number of hard negatives to 8. Results are shown in Tab.~\ref{tab:sd_effectiveness}.
We compare six settings: (1) random hard-negative sampling with hard one-hot weights $W_c=\{1,0,\dots,0\}$; (2) EVA-CLIP-based hard-negative sampling with hard one-hot weights; (3) SD-guided hard-negative sampling with hard one-hot weights; (4) SD-guided sampling with 30\% inverted SD weights, where $W_{c_i}\leftarrow 1-W_{c_i}$ for 30\% of sampled negatives; (5) SD-guided sampling with 10\% inverted SD weights; and (6) our full method using SD-guided sampling with the original SD weights.

Tab.~\ref{tab:sd_effectiveness} yields several observations. First, the full SD-based setting performs best across benchmarks, showing that combining SD-guided hard-negative construction with soft supervision is most effective. Second, under identical hard one-hot supervision, performance improves from random to CLIP-based and then SD-guided sampling, demonstrating that SD better identifies informative and semantically confusable negatives. Third, replacing SD weights with hard one-hot targets causes a drop even with the same SD-based negatives, confirming that the gains stem not only from better negative construction but also from soft supervision. Finally, corrupting SD weights consistently degrades performance: the 10\% inverted setting underperforms the full method, while the 30\% setting drops further and even falls below the SD hard-one-hot variant. This indicates that SD supervision is effective because it captures semantic relations among candidates. Overall, these results validate the effectiveness of SD supervision for retrieval learning.

\begin{table}[t]
\caption{Analysis of different hard-negative construction and supervision strategies for studying the effectiveness of SD signals. All settings use 8 hard negatives per query.}
\vspace{-2mm}
\label{tab:sd_effectiveness}
\centering
\setlength{\tabcolsep}{3.8pt}
\renewcommand{\arraystretch}{1.12}
\resizebox{\columnwidth}{!}{
\begin{tabular}{ll | cc | cc}
\toprule
\multirow{2.5}{*}{\textbf{Negative Sampling}} & \multirow{2.5}{*}{\textbf{Training Weights}} & \multicolumn{2}{c|}{\textbf{E-VQA}} & \multicolumn{2}{c}{\textbf{InfoSeek}} \\
\cmidrule(lr){3-4} \cmidrule(lr){5-6}
& & R@1 & R@5 & R@1 & R@5 \\
\midrule
Random & Hard one-hot weights                      & 9.1  & 18.7 & 20.3 & 36.8 \\
CLIP similarity & Hard one-hot weights             & 14.7 & 34.9 & 47.1 & 70.3 \\
SD & Hard one-hot weights                          & 16.4 & 40.1 & 53.1 & 71.6 \\
SD & 30\% inverted SD weights                      & 12.1 & 30.4 & 43.3 & 62.1 \\
SD & 10\% inverted SD weights                      & 19.9 & 38.2 & 54.4 & 68.8 \\
\rowcolor{cyan!15}
SD & SD weights (ours)                             & \textbf{24.7} & \textbf{48.7} & \textbf{60.3} & \textbf{74.7} \\
\bottomrule
\end{tabular}
}
\vspace{-2mm}
\end{table}

\textbf{Do SD supervision signals faithfully capture entity-aware semantics?}
To examine whether SD supervision faithfully captures entity-aware semantics, we construct a pairwise evaluation set using entity labels from the dataset, including 10,000 positive pairs (two images of the same entity) and 10,000 negative pairs (two images of different entities). For each pair, we compute two scores: (1) EVA-CLIP cosine similarity, linearly rescaled from $[-1,1]$ to $[0,1]$ via $\hat{s}=(s+1)/2$, and (2) the SD semantic consistency weight. We use AUC as the primary measure of discriminability.

The results show that both CLIP similarity and SD weight assign higher scores to positive pairs than to negative pairs, indicating that both carry useful entity-related information. However, SD shows a substantially larger positive--negative separation and achieves a clearly higher AUC than CLIP similarity (0.91 vs. 0.79). This suggests that SD is more faithful to true entity identity, while CLIP similarity mainly reflects surface-level visual resemblance.
Overall, these results provide direct evidence that SD supervision signals faithfully capture entity-aware semantics by encoding semantic relations that are more closely aligned with true entity identity.

\begin{table}[ht]
\centering
\vspace{-2mm}
\caption{Ablation study on different Semantic Discriminator.}
\vspace{-2mm}
\begin{tabular}{c|cc c cc}
\toprule
    \multirow{2.5}{*}{SD} &  \multicolumn{2}{c}{\textbf{E-VQA}} & & \multicolumn{2}{c}{\textbf{InfoSeek}} \\
    \cmidrule{2-3} \cmidrule{5-6}
    & R@1 & R@5 &  & R@1 & R@5\\
    \midrule

 InternVL3-8B  & 20.7 & 44.3 & & 54.4 & 67.9 \\
 InternVL3-14B  & 24.4 & \textbf{49.1} & & 59.1 & 73.4\\
  \rowcolor{cyan!15}
 Qwen2.5-VL-7B  & \textbf{24.7} & 48.7 & & \textbf{60.3} & \textbf{74.7}  \\
\bottomrule
\end{tabular}
\label{tab:ablation_SD}
\vspace{-2mm}
\end{table}

\textbf{Ablation on Different Semantic Discriminator.}
Since the semantic discriminator is responsible for producing entity consistency weights, its comprehension capability directly affects the supervision quality and thus the final retrieval performance. 
Therefore, we compare two influential MLLMs in the current open-source community: Qwen2.5-VL-7B, InternVL3-8B, and InternVL3-14B. 
Tab.~\ref{tab:ablation_SD} shows that Qwen2.5-VL-7B yields the strongest results, substantially outperforming InternVL3-8B on both E-VQA and InfoSeek. Scaling InternVL3 from 8B to 14B brings clear gains, yet it remains slightly behind Qwen2.5-VL-7B.

\begin{table}[ht]
\centering
\vspace{-2mm}
\caption{Ablation study on the number of hard negatives.}
\vspace{-2mm}
\begin{tabular}{c|cc c cc}
\toprule
    \multirow{2.5}{*}{\#Negatives} &  \multicolumn{2}{c}{\textbf{E-VQA}} & & \multicolumn{2}{c}{\textbf{InfoSeek}} \\
    \cmidrule{2-3} \cmidrule{5-6}
    & R@1 & R@5 &  & R@1 & R@5\\
    \midrule
 4  & 22.5                             & 45.1 &                    & 57.7                               & 69.8  \\
 6  & 23.3                             & 46.8 &                    & 59.1                               & 72.4  \\
\rowcolor{cyan!15}
 8  & \textbf{24.7}                             & \textbf{48.7} &                    & \textbf{60.3}   & 74.7  \\
10  & 24.2 & \textbf{48.7} & & 60.2                               & \textbf{74.8}  \\
\bottomrule

\end{tabular}
\label{tab:ablation_neg_num}
\vspace{-2mm}
\end{table}

\begin{table}[t]
\centering
\caption{Ablation on the semantic sharpness coefficient $\gamma$.}
\vspace{-2mm}
\begin{tabular}{c|c|cc c cc}
\toprule
    \multirow{2.5}{*}{Model} &\multirow{2.5}{*}{$\gamma$} &  \multicolumn{2}{c}{\textbf{E-VQA}} & & \multicolumn{2}{c}{\textbf{InfoSeek}} \\
    \cmidrule{3-4} \cmidrule{6-7}
    & & R@1 & R@5 &  & R@1 & R@5\\
    \midrule
\multirow{4}{*}{KBMR (LLaVA-OV-7B)} 
 & 0.9  & 23.3             & 46.0 &          & 57.6          & 71.2  \\
 & 1.0  & 24.1             & 47.8 &          & 59.1          & 72.4  \\
 & \cellcolor{cyan!15}1.1  & \cellcolor{cyan!15}\textbf{24.7}    & \cellcolor{cyan!15}\textbf{48.7} &  \cellcolor{cyan!15}    & \cellcolor{cyan!15}\textbf{60.3}   & \cellcolor{cyan!15}\textbf{74.7}  \\
 & 1.2  & 24.4 & 48.0 & & 59.4                               & 73.2  \\
\bottomrule
\end{tabular}
\label{tab:ablation_gama}
\vspace{-2mm}
\end{table}

\textbf{Ablation on the Number of Hard Negatives.}
Tab.~\ref{tab:ablation_neg_num} analyzes how the number of mined hard negatives affects retrieval performance based on KBMR(LLaVA-OV-7B). Increasing Negatives from 4 to 8 consistently boosts results on both E-VQA and InfoSeek, suggesting that richer hard-negative supervision strengthens the model’s ability to separate highly confusable candidates. 
Meanwhile, increasing the number further to 10 does not lead to a significant performance gain.

\textbf{Effect of the semantic sharpness coefficient $\gamma$.}
Tab.~\ref{tab:ablation_gama} studies the semantic sharpness coefficient $\gamma$, which controls the temperature for converting semantic discriminator logits into continuous consistency weights. On both E-VQA and InfoSeek, performance improves from $\gamma{=}0.9$ to $\gamma{=}1.1$, where the best results are achieved. Small $\gamma$ makes the weights overly sharp and close to hard binary labels, while large $\gamma$ over-smooths the distribution and weakens the contrast among candidates. These results suggest that $\gamma{=}1.1$ provides a better balance between discrimination and smoothness, so we use it in all experiments.

\vspace{-1mm}
\begin{figure}[ht]
\centering
\includegraphics[width=\linewidth]{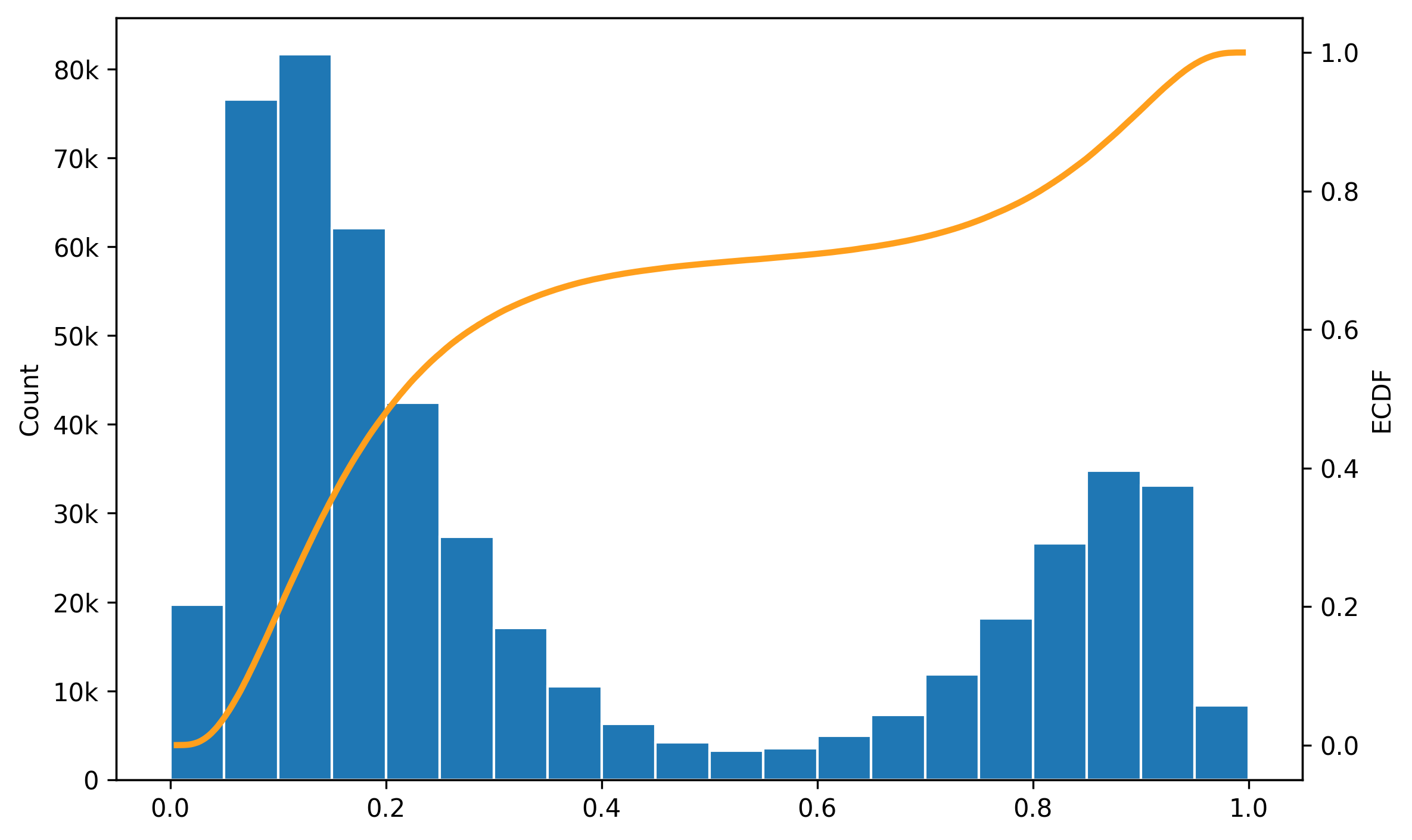}
\vspace{-6mm}
\caption{Global distribution of semantic consistency weights across the Top-50 highest-similarity negative candidates.}
\vspace{-2mm}
\label{fig:weight}
\end{figure}

\textbf{Weight Distribution Analysis.}
We analyze the global distribution of semantic consistency weights $w$ by randomly sampling 10,000 training queries and collecting the Top-50 most similar candidates (the potential hard negative set) for each query. Although these candidates are retrieved by a CLIP-style similarity metric and are thus visually competitive, many receive low weights, as shown in Fig.~\ref{fig:weight}, indicating that the MLLM evaluation can identify semantic mismatch even within confusing neighborhoods. This confirms that \textbf{MLLM-based semantic supervision offers stronger discrimination than CLIP's visual similarity alone}.
At the same time, the existence of high-weight candidates reflects challenging or ambiguous cases that remain difficult even for MLLMs. Importantly, the weight distribution is not purely binary: a considerable mass lies in the intermediate range, reflecting \textbf{graded difficulty among hard negatives} and providing \textbf{continuous soft supervision} with more stable learning signals than 0/1 labels.

\begin{figure}[!t]
\centering
\includegraphics[width=\linewidth]{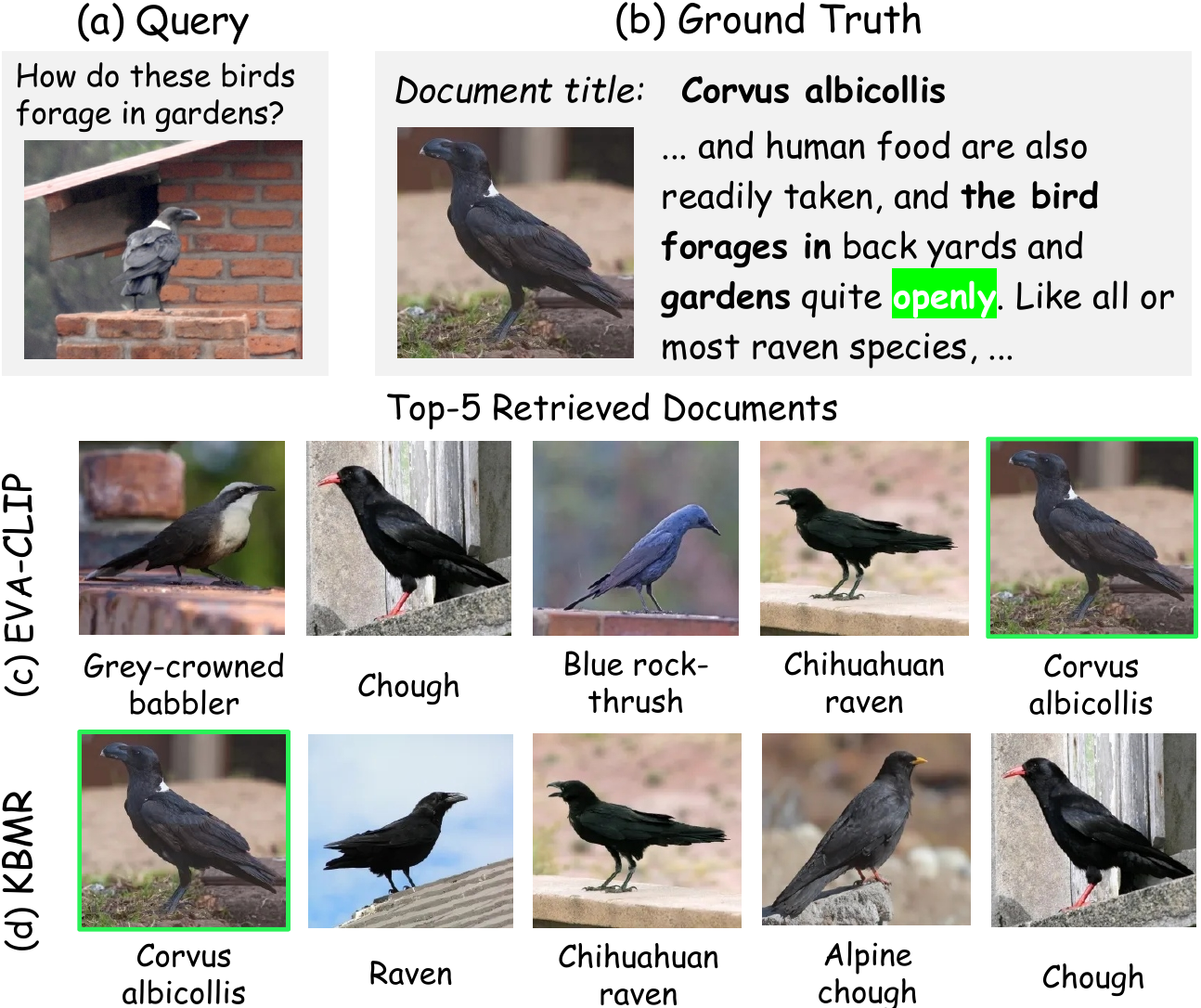}
\vspace{-4mm}
\caption{\textbf{Qualitative analysis on KB-VQA retrieval}. (a) shows the query with an image of a Corvus albicollis; (b) shows the GT document of Corvus albicollis; (c) and (d) show the top-5 retrieved images from EVA-CLIP and our proposed KBMR, respectively. The answer is highlighted in green.}
\vspace{-4mm}
\label{fig:result}
\end{figure}

\textbf{Visualization Analysis Results.}
Fig.~\ref{fig:result} provides a qualitative comparison of KB-VQA retrieval. We visualize the top-5 retrieved results from EVA-CLIP and our KBMR. EVA-CLIP returns several visually similar but semantically mismatched bird species and ranks the correct entity (Corvus albicollis) only at the 5th position. In contrast, KBMR places the ground-truth document at rank-1 and retrieves a more semantically coherent neighborhood. This example suggests that KBMR better aligns retrieval with entity-level semantics rather than surface-level visual similarity, leading to stronger evidence recall for downstream KB-VQA.

\section{Conclusion}
We introduced KBMR, the first MLLM-based retriever for knowledge-based visual question answering. KBMR addresses the fundamental mismatch in CLIP-style retrieval, where surface-level visual similarity is often an unreliable proxy for entity-level semantic relevance under large appearance variations. By leveraging autoregressive semantic embeddings from MLLMs and an MLLM-based semantic discriminator that provides continuous entity-consistency weights, KBMR enables reliable hard-negative construction and semantically informed soft supervision for Wikipedia-scale retrieval. We further proposed a continuous semantic distillation objective that aligns the retriever-induced similarity distribution with a discriminator-induced semantic prior, encouraging fine-grained, entity-centric discrimination in highly confusable neighborhoods.
Extensive experiments on E-VQA, InfoSeek, and OK-VQA demonstrate that KBMR consistently improves retrieval recall and delivers substantial gains in end-to-end KB-VQA performance. More importantly, our work shows that KB-VQA retrieval should not be limited to surface-level visual matching, but should explicitly model entity-aware semantic relations. By doing so, KBMR alleviates a key bottleneck in KB-VQA pipelines, providing both a stronger retriever and a higher-quality candidate pool for downstream reasoning and answer grounding in practical KB-VQA systems.

\section{Acknowledgements}
This work is supported by the NSFC fund (62576190), in part by the Shenzhen Science and Technology Project under Grant (KJZD2024\allowbreak0903103210014)

\bibliographystyle{ACM-Reference-Format}
\balance
\bibliography{reference}

\end{document}